\documentclass[letterpaper]{article} %
\usepackage{aaai24}  %
\usepackage{times}  %
\usepackage{helvet}  %
\usepackage{courier}  %
\usepackage[hyphens]{url}  %
\usepackage{graphicx} %
\usepackage{natbib}  %
\usepackage{caption} %
\usepackage{algorithm}
\usepackage{algorithmic}

\usepackage{newfloat}
\usepackage{listings}
\DeclareCaptionStyle{ruled}{labelfont=normalfont,labelsep=colon,strut=off} %
\floatstyle{ruled}
\newfloat{listing}{tb}{lst}{}
\floatname{listing}{Listing}
\title{\tool{}: Automated Simulations of Adversarial Attacks on \\Arbitrary Objects in Realistic Scenes}
\author {
    Matthew Hull\textsuperscript{\rm 1},
    Zijie J. Wang\textsuperscript{\rm 1},
    Duen Horng (Polo) Chau\textsuperscript{\rm 1}
}
\affiliations {
    \textsuperscript{\rm 1}Georgia Institute of Technology\\
    \{matthewhull, jayw, polo\}@gatech.edu
}
\usepackage{bibentry}
\usepackage{enumitem}
\usepackage{rotating}
\usepackage{tikz}
\usepackage{capt-of,etoolbox}
\usepackage{microtype}
\usepackage{bm}
\usepackage{soul}
\usepackage{graphbox}
\usepackage{tcolorbox}
\usepackage{mathtools}
\usepackage[varqu]{zi4}
\usepackage[bb=boondox]{mathalfa}
\usepackage{units}
\usepackage{gensymb}
\usepackage{printlen} %
\usepackage{placeins} %
\usepackage{afterpage} %
\usepackage{amsmath}
\usepackage{eqnarray}
\definecolor{orange}{RGB}{250,130,49}
\definecolor{red}{RGB}{234,59,90}
\definecolor{agreen}{RGB}{74, 198, 148}
\definecolor{purple}{RGB}{158, 62, 177}
\definecolor{darkpurple}{RGB}{170, 70, 210}
\definecolor{aqua}{RGB}{87, 180, 181}
\definecolor{lightblue}{RGB}{72, 123, 232}
\definecolor{hotpink}{RGB}{255, 83, 115}
\definecolor{teal}{RGB}{90, 200, 250}
\definecolor{linkColor}{RGB}{0, 128, 229}
\definecolor{lightgreen}{RGB}{33, 222, 128}
\definecolor{gray}{RGB}{75, 101, 132}

\newcommand{\tool}{\textsc{\textsf{Revamp}}}
\newcommand{\github}{\url{https://github.com/poloclub/revamp}}
\newcommand{\persona}{Jim}

\usepackage{xcolor}
\definecolor{codebg}{gray}{0.96}
\definecolor{codeframe}{gray}{0.8}
\definecolor{stop_sign_red}{RGB}{189,44,38}
\definecolor{mailbox_blue}{RGB}{47, 97, 216}
\definecolor{multicam_green}{RGB}{65, 146, 49}
\definecolor{code_gray}{RGB}{115, 115, 115}

\newtoggle{inheader}
\begin{document}

\maketitle

\begin{abstract}
\label{sec:abstract}
Deep Learning models, such as those used in an autonomous vehicle are vulnerable to adversarial attacks where an attacker could place an adversarial object in the environment, leading to mis-classification.
Generating these adversarial objects in the digital space has been extensively studied, however successfully transferring these attacks from the digital realm to the physical realm has proven challenging when controlling for real-world environmental factors. 
In response to these limitations, we introduce \tool{}, an easy-to-use Python library that is the first-of-its-kind tool for creating attack scenarios with arbitrary objects and simulating realistic environmental factors, lighting, reflection, and refraction. 
\tool{} enables researchers and practitioners to swiftly explore various scenarios within the digital realm by offering a wide range of configurable options for designing experiments and using differentiable rendering to reproduce physically plausible adversarial objects.
We will demonstrate and invite the audience to try \tool{} to produce an adversarial texture on a chosen object while having control over various scene parameters. 
The audience will choose a scene, an object to attack, the desired attack class, and the number of camera positions to use. Then, in real time, we show how this altered texture causes the chosen object to be mis-classified, showcasing the potential of \tool{} in real-world scenarios. 
\tool{} is open-source and available at \url{https://github.com/poloclub/revamp}. \end{abstract}

\section{Introduction}
\label{sec:introduction}
\begin{figure}[t]
\centering
\includegraphics[width=1.0\columnwidth]{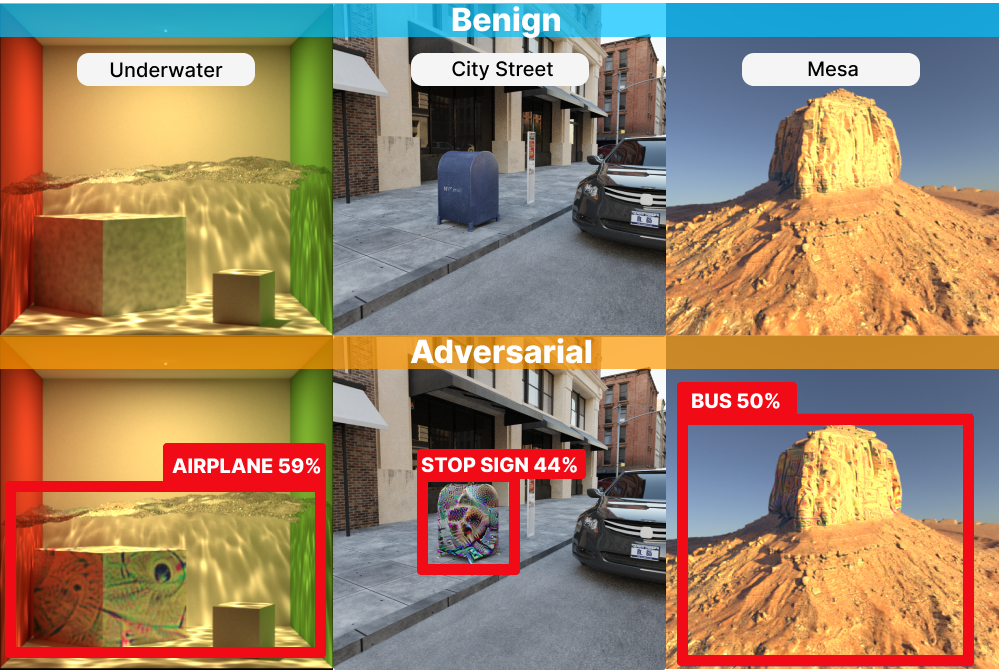}
\caption{\tool{} allows user to easily perturb textures on arbitrary objects with a variety  lighting and environmental settings in realistic scenes.
\textbf{Top Row:} Benign examples for 3 different scenes. 
\textbf{Bottom Row:} Perturbed texture maps on an underwater cube with light distortions perturbed as an airplane, a mailbox on a city street attacked to appear as a stop sign, and a mesa attacked to appear as a bus.}
\label{fig1}
\end{figure}
\begin{figure*}[t]
\centering
\includegraphics[width=0.9\textwidth]{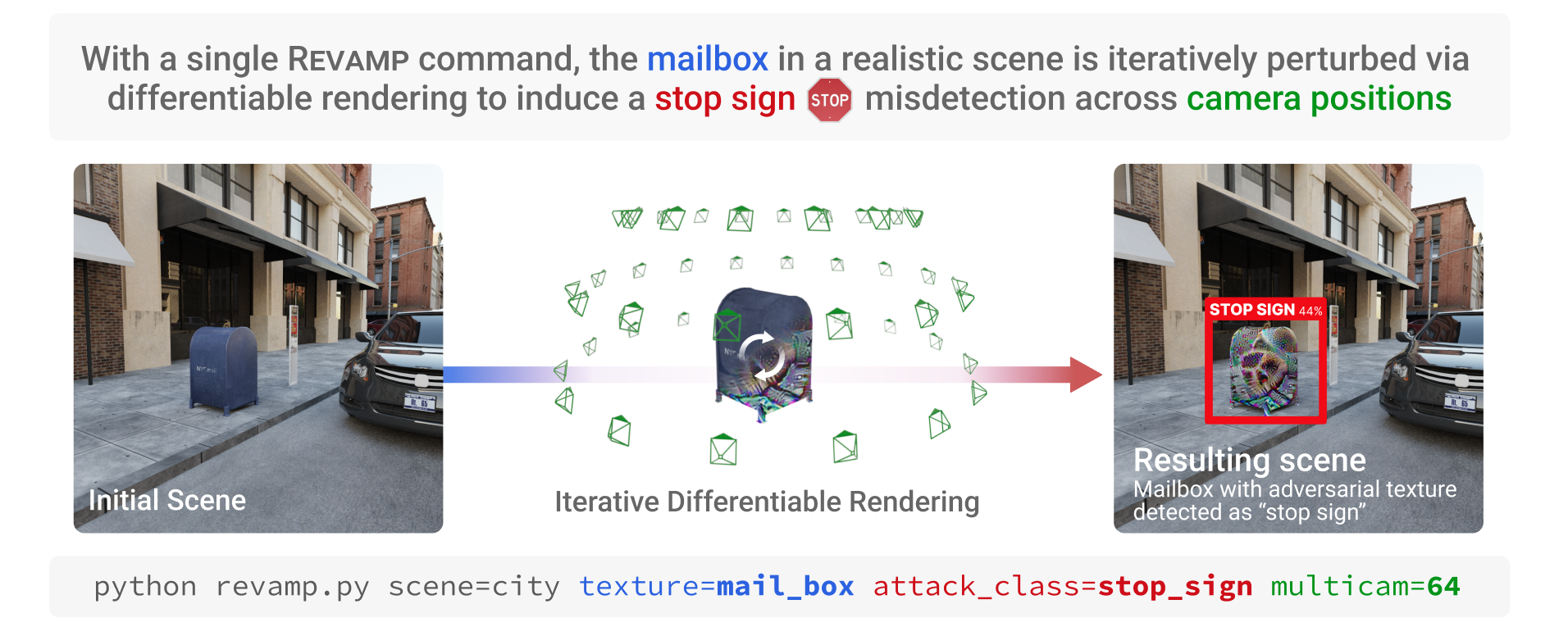} 
\caption{\tool{}'s pipeline can be run with a simple command, allowing a user to choose a scene, an object to attack, the desired attack class, and the number of camera positions to use.  Differentiable rendering is used to iteratively render and update the texture of the attacked object.}
\label{fig2}
\end{figure*}
It is widely recognized that deep neural networks employed in image recognition and object detection tasks are vulnerable to adversarial attacks \cite{goodfellow2015harnessadv}.
While these attacks were initially produced in the digital space, there is growing interest in exploring the feasibility of physically realizable adversarial attacks \cite{chenshapeshifter}.
Subsequent work explored various methods to create physical adversarial objects, including printing adversarial textures onto objects \cite{chenshapeshifter}, projection mapping \cite{huang2022spaa}, and marker and sticker placement \cite{song2018physadv}. 
However, transferring successful attacks from the digital realm to the physical realm
 remains challenging. 
Techniques like Expectation over Transformation (EoT) \cite{athalye2018eot} focus mainly on limited scaling and rotation transformations, often neglecting real-world factors such as environmental conditions and lighting. 

Experimentation in 3D environments that realistically portray the real world has strong potential to address the these limitations \cite{xiao2019meshadv}.
However,
creating such environments require expertise in 3D modeling and rendering.
Furthermore, traditional 3D rendering lacks support for optimizing scene parameters and textures due to non-differentiable operations. 

To address these challenges,  our research 
 make the following major contributions:

\begin{itemize}

    \item \textbf{We introduce \tool{}, the first-of-its kind open-source Python library for easily creating 
    attack scenarios with arbitrary objects in realistic scenes.}
    \tool{} empowers researchers and practitioners to swiftly explore various attack scenarios, offering a wide range of configurable options for designing experiments in a customizable adversarial attack pipeline for creating 3D objects. 
    \tool{} uses \textit{differentiable rendering}, which propagates gradients through the rendering to optimize the scene parameters to induce successful attacks \cite{mitsuba3}.
With \tool{}, users can select from scenes that replicate real-world scenarios, or create their own. 
They can designate textures and parameters to craft tailored adversarial examples. 
Our library of scenes support realistic lighting effects like reflection and refraction, and other realistic environmental factors (Fig.\ref{fig1}).
\tool{} stands for \textit{\textbf{R}eshaping \textbf{E}nvironment by \textbf{V}arying \textbf{A}dversarial \textbf{M}odel's \textbf{P}arameters}.

    \item \textbf{Open-Source Implementation:} 
    While some prior works \cite{liu2018beyond, xiao2019meshadv} have explored differentiable rendering for adversarial examples, they are closed-source.
    \tool{} is open-source,\footnote{\tool{} code: \github{}} facilitating collaboration and reproducible research for machine learning (ML) security researchers and practitioners.

\end{itemize}

\section{System Design and Implementation}
\label{sec:sys-design-implementation}
 \tool{} uses the the Hydra configuration management tool \cite{Yadan2019Hydra} to construct and customize scenarios, providing flexibility by allowing parameter adjustments through the command line. 
Each scenario is provided as a \texttt{YAML} configuration file including a 3D scene, an attackable parameter, render settings, and a victim model. 
For example, in a ``city" scenario, a camera is focused on a mailbox, with its texture as the ``attackable" parameter~(Fig.\ref{fig2}).
The scenario configuration specifies the inference model, optimized parameter(s), and attack type (targeted or untargeted). 
Camera transformations, such as orbiting, can also be configured, offering diverse perspectives for exploring adversarial impacts.
Importantly, \tool{} employs the Mitsuba differentiable renderer \cite{mitsuba3} with ray tracing to precisely simulate real-world phenomena like reflection, shadows, and refraction. This enhances scene realism and accuracy, enabling more precise evaluation of adversarial attacks~(Fig.\ref{fig2}). 

\section{Usage Scenario}
\label{sec:usage-scenario}
\persona{} is an ML security engineer at a self-driving car company. He wants to leverage \tool{} to assess and enhance the object detector's robustness for potential deployment in self-driving cars.
Interested in investigating how a large object in a street environment might be misclassified \textit{e.g.}, a stop sign, if manipulated adversarially, \persona{} selects a city scene and designates the mailbox's texture as the attackable parameter. 

To conduct this simulation, \persona{} executes \tool{}'s pipeline with a simple command (Fig.\ref{fig2}). 
At each step of differentiable rendering, the resulting scene image is forwarded to the object detector, and the texture is iteratively perturbed using a PGD $\ell_2$ attack \cite{madry2018towards} until a ``stop sign'' detection is achieved. 
\tool{} outputs rendered images, object detection bounding boxes, class labels, and perturbed texture maps for each rendering step. 
\persona{} notes substantial perturbation was required before a ``stop sign'' detection with low confidence was achieved.

Following the pipeline run, \persona{} has gained confidence that the object detector is resistant 
against a strong perturbation, and proceeds to use the output texture maps to conduct adversarial training to further strengthen model robustness.
The ease of use of \tool{} also encourages \persona{} to quickly expand his experiment to more realistic scenes such as open highways or unmarked roads in wooded areas.

\section{Conclusion}
\label{sec:conclusion}
We present \tool{}, an open-source, easy-to-use Python library that helps ML researchers create attack scenarios with arbitrary objects while simulating environmental factors and lighting with differentiable rendering. 
\tool{} helps users explore these scenarios in the digital realm to help close the gap between digital and physically plausible attacks.

\bibliography{aaai24}

\end{document}